\documentclass[letterpaper]{article} 
\usepackage{aaai2026}  
\usepackage{times}  
\usepackage{helvet}  
\usepackage{courier}  
\usepackage[hyphens]{url}  
\usepackage{graphicx} 
\usepackage{natbib}  
\usepackage{caption} 
\usepackage{algorithm}
\usepackage{algorithmic}

\usepackage{newfloat}
\usepackage{listings}
\DeclareCaptionStyle{ruled}{labelfont=normalfont,labelsep=colon,strut=off} 
\floatstyle{ruled}
\newfloat{listing}{tb}{lst}{}
\floatname{listing}{Listing}
\usepackage{amsfonts}
\usepackage{amsmath}
\usepackage{amssymb}
\usepackage{mathrsfs}
\usepackage{booktabs} 
\usepackage{multirow} 
\usepackage{xcolor}
\usepackage{float}
\usepackage{graphicx}
\usepackage{subfig}
\usepackage{threeparttable}
\usepackage{pifont}
\usepackage{bbding}
\usepackage{colortbl}
\usepackage{enumitem}

\newtheorem{theorem}{Theorem}

\newtheorem{lemma}{Lemma}

\newcommand{\M}{TOFA}

\title{TOFA: Training-Free One-Shot Federated Adaptation for \\Vision-Language Models}

\author{
    Li Zhang\textsuperscript{\rm 1},
    Zhongxuan Han\textsuperscript{\rm 1},
    XiaoHua Feng\textsuperscript{\rm 1},
    Jiaming Zhang\textsuperscript{\rm 1},
    Yuyuan Li\textsuperscript{\rm 2}\thanks{Yuyuan Li is the corresponding author.},
    Linbo Jiang\textsuperscript{\rm 3},\\
    Jianan Lin\textsuperscript{\rm 3},
    Chaochao Chen\textsuperscript{\rm 1}
}
\affiliations{
    \textsuperscript{\rm 1}College of Computer Science and Technology, Zhejiang University\\
    \textsuperscript{\rm 2}School of Communication Engineering, Hangzhou Dianzi University\\
    \textsuperscript{\rm 3}Ant Group\\

    zhanglizl80@gmail.com, \{zxhan, fengxiaohua, 22321350\}@zju.edu.cn, y2li@hdu.edu.cn\\
    xiaobo.jlb@antgroup.com, jianan.linjn@myxiaojin.cn, zjuccc@zju.edu.cn

}

\usepackage{bibentry}

\begin{document}

\maketitle
\begin{abstract}
Efficient and lightweight adaptation of pre-trained Vision-Language Models (VLMs) to downstream tasks through collaborative interactions between local clients and a central server is a rapidly emerging research topic in federated learning.
Existing adaptation algorithms are typically trained iteratively, which incur significant communication costs and increase the susceptibility to potential attacks.
Motivated by the one-shot federated training techniques that reduce client-server exchanges to a single round, developing a lightweight one-shot federated VLM adaptation method to alleviate these issues is particularly attractive.
However, current one-shot approaches face certain challenges in adapting VLMs within federated settings: (1)~\textit{insufficient exploitation of the rich multimodal information inherent in VLMs}; (2)~\textit{lack of specialized adaptation strategies to systematically handle the severe data heterogeneity}; and (3) \textit{requiring additional training resource of clients or server}.
To bridge these gaps, we propose a novel \textbf{T}raining-free \textbf{O}ne-shot \textbf{F}ederated \textbf{A}daptation framework for VLMs, named~\M. 
To fully leverage the generalizable multimodal features in pre-trained VLMs, TOFA employs both visual and textual pipelines to extract task-relevant representations.
In the visual pipeline, a hierarchical Bayesian model learns \textbf{personalized}, class-specific prototype distributions. For the textual pipeline, \M~evaluates and globally aligns the generated local text prompts for \textbf{robustness}. An adaptive weight calibration mechanism is also introduced to combine predictions from both modalities, balancing personalization and robustness to handle data heterogeneity.
Our method is training-free, not relying on additional training resources on either the client or server side.
Extensive experiments across 9 datasets in various federated settings demonstrate the effectiveness of the proposed \M~method.
\end{abstract}


\section{Introduction}
Federated learning (FL)~\cite{fedavg}, a distributed machine learning paradigm, enables multiple clients to collaboratively refine a shared model while preserving their data privacy. 
Recent advancements in large-scale pre-trained models, particularly Vision-Language Models (VLMs) such as CLIP~\cite{CLIP} and ALIGN~\cite{ALIGN-pmlr-v139-jia21b}, have gained widespread attention within FL, driven by their impressive capability to learn transferable representations.
An increasing number of contemporary studies focus on the adaptation of VLMs to improve performance in downstream tasks, particularly using fine-tuning and prompt learning~\cite{fed-vlm-finetuning-zhang2025enhancing,ref-6,ref-5,portfolio-fed-theory-pan2024federated,ref-3,ref-4,ref-9,ref-8,fedpgp-harmon-pmlr-v235-cui24c,ref-1,ref-7,ref-2,text-driven-prompt-fed-qiu2024federated}.
However, the large parameter sizes inherent to VLMs result in substantial communication and computation overhead, significantly hindering their practical applicability in FL scenarios.
Most federated VLM adaptation methods, whether fine-tuning or prompt learning, heavily rely on multi-round interactions between the server and clients, increasing communication burdens and requiring sustained system robustness and reliability~\cite{fed-servey-open-problem-kairouz2021advances,oneshot-fedlpa-liu2024fedlpa}.
%
%
%
Moreover, a large number of clients and servers (e.g., most mobile devices and capability‑limited servers) lack the computational resources necessary for model training~\cite{fed-mobile-abreha2022federated,fed-privacy-server-bonawitz2019towards,fed-Opportunities-challenges-mammen2021federated}, thereby impeding the deployment of VLMs in distributed settings.
Recently, the one-shot FL technique emerges as an effective method to minimize communication overhead by consolidating client-server interactions into a single round~\cite{oneshot-fedlpa-liu2024fedlpa,oneshot-fens-allouah2024revisiting,causal-fed-one-shot-tang2024fusefl,synthetic-distill-oneshot-zhang2024one}, significantly reducing communication overhead while preserving privacy.
Motivated by this, it is compelling to formulate a \textbf{lightweight one-shot federated adaptation framework for VLMs} to address the aforementioned issues.
%

Despite the promising advancements of one-shot techniques, developing both training-free and one-shot federated VLMs adaptation methods still faces certain challenges:
(1) \textit{Insufficient exploitation of the rich multimodal information inherent in VLMs.} 
Most one-shot training frameworks are designed for traditional federated model training~\cite{oneshot-fedlpa-liu2024fedlpa,oneshot-fens-allouah2024revisiting,causal-fed-one-shot-tang2024fusefl,synthetic-distill-oneshot-zhang2024one}, primarily underlining the visual modality and lacking the capacity to leverage the rich multimodal information from VLMs.
%
%
%
These methods are ill-suited for efficiently adapting pre-trained VLMs to the FL setting, as they fail to capture the critical interactions between visual and textual modalities.
%
%
(2) \textit{Lack of tailored adaptation strategies to effectively handle severe data heterogeneity.}
The data heterogeneity within decentralized training frameworks further impedes the development of lightweight adaptation methods for VLMs in FL, where the data distributions among clients are non-identically and non-independently distributed (non-IID)~\cite{fedprox-li2020federated,fedBN-li2021fedbn,Liu-ref-2,liu-ref-1}. 
Such non-IID nature causes distributional shifts from client data to the global data, leading to discrepancies in local and global optimization objectives, which prevent existing VLM adaptation methods from meeting clients' personalized demands, underscoring the necessity for adaptation strategies tailored to federated settings.
%
%
%
(3) \textit{Additional training resource requirements for the clients or server.}
Existing VLMs adaptation methods within FL typically depend on additional model training resources on the client or server side~\cite{feddisc-Yang_Su_Li_Xue_2024,mixexport-luo2025mixtureexport,DP-FPL-tran2025privacypreserving}.
These methods are incompatible in the ubiquitous resource-constrained distributed environments~\cite{fed-Opportunities-challenges-mammen2021federated}.
%
%

%

In this paper, we introduce a novel training-free one-shot federated adaptation approach for VLMs to bridge these gaps, named \M.
%
\textbf{To fully leverage the generalizable multimodal features within VLMs}, our approach employs both visual and textual pipelines to extract task-relevant representations for downstream classification.
For the visual pipeline, we utilize a hierarchical Bayesian model to model the heterogeneous feature representations of class-specific prototypes derived from the visual encoder, with the global information serving as the prior for the inference of local feature distributions.
The classification probability for downstream tasks is derived using Gaussian Discriminant Analysis (GDA) on the posterior prompt distributions of class prototypes.
%
%
For the textual pipeline, \M~aims to extract robust and generalizable augmented textual inputs from the textual modality. 
The raw text inputs in the original CLIP-based classification are first augmented using Large Language Models. 
These inputs are then evaluated for quality on each client and globally aligned to select robust text prompts that consistently show high importance scores across diverse local environments.
\textbf{To address the impact of data heterogeneity}, our approach fuses personalized visual representations with robust text augmentations to integrate both global and local information, learning locally adaptive features while preventing the model from overfitting.
Specifically, \M~introduces an adaptive weight calibration technique that adjusts the sample-wise contributions of the textual and visual modalities based on their prediction confidence to strike a balance between personalization and generalization.
Furthermore, throughout the training process, our approach operates \textbf{without relying on training resources on either the server or client side}, thereby enhancing its practicality and flexibility.
%
%
%

We demonstrate the effectiveness of \M~by conducting experiments across nine datasets containing vision datasets and domain datasets within various data heterogeneous environments.
%
%
\M~has a consistent and significant improvement over existing one-shot baselines, and even surpasses several training-based federated VLM adaptation methods.

\noindent Our main contributions can be summarized as follows:
\begin{itemize}
    \item To our best of knowledge, we are the first to propose an effective training-free one-shot adaptation method for VLMs in the FL setting. 
    \item We propose an one-shot visual pipeline learn the personalized class-specific prompt distribution over visual representation and a textual pipeline to extract robust text augmentations though global text alignment.
    \item We propose a sample-adaptive modality weight calibration method to integrate personalized visual representations and robust text representations, allowing the model to handle data heterogeneity within FL.
    \item We conducted extensive experiments on widely adopted datasets in various data heterogeneity, and significant result improvement verifies the superiority of~\M.
\end{itemize}



\section{Preliminary}
\begin{figure*}[t]
\centering
\includegraphics[width=0.95\linewidth]{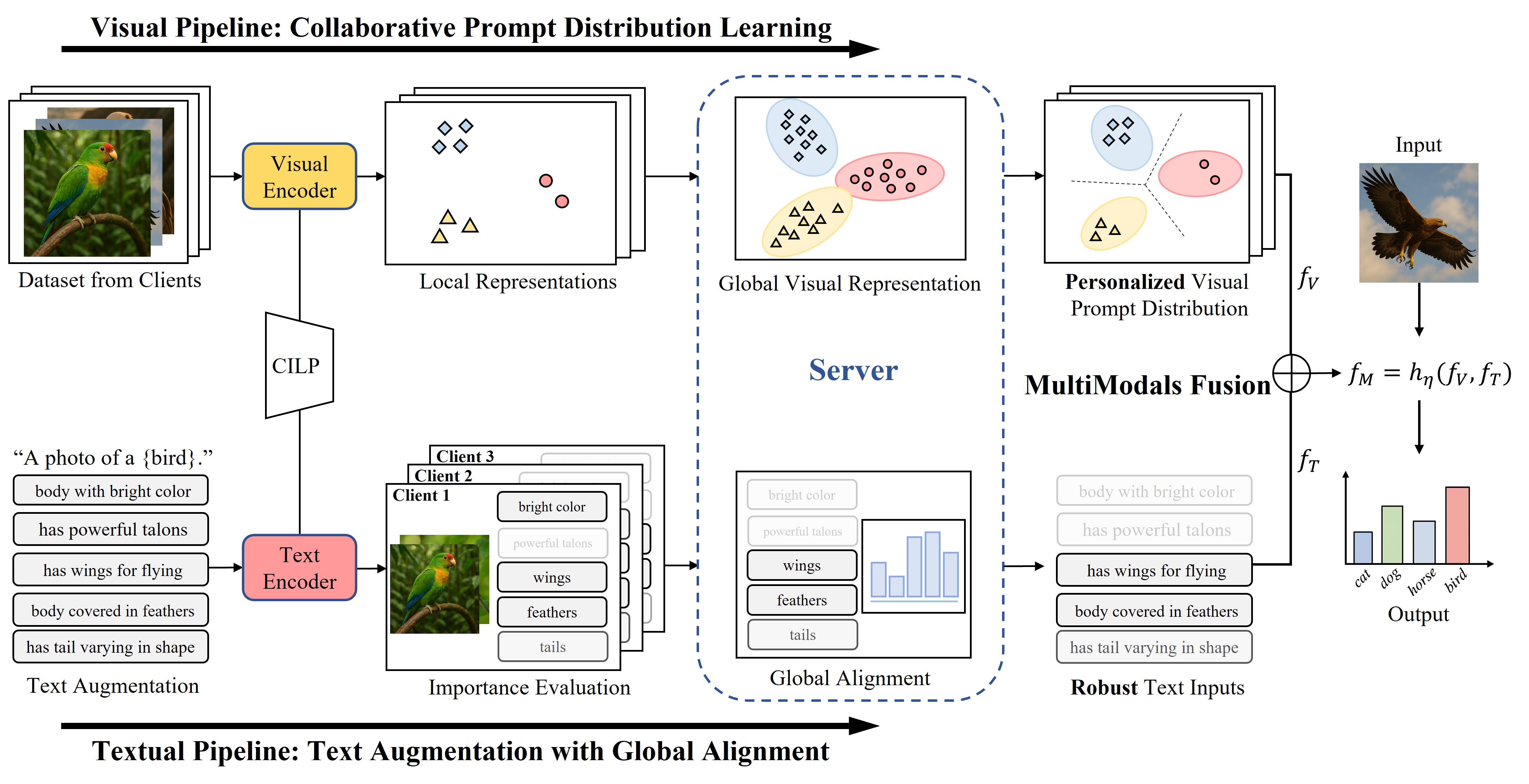} 
\vspace{-5pt}
\caption{Overall Framework of~\M}
\vspace{-5pt}
\label{framework-fig-1}
\end{figure*}

In this section, we focus on the background of VLMs and FL. Detailed related work is presented in \textbf{Appendix A}.

\noindent \textbf{Contrastive Language-Image Pretraining (CLIP).} CLIP \cite{CLIP} consists of a visual encoder $\Phi_\mathbf{V}(\mathbf{x})$ and text encoder $\Phi_\mathbf{T}(\mathbf{t})$, each producing a normalized $d$-dimensional embedding from an arbitrary image $\mathbf{x}$, and word embeddings $\mathbf{t}$.
%
%
Once trained, CLIP enables zero-shot $C$-class image classification by generating each of the $c$ classifier weights $\mathbf{w}_c$ as the $d$-dimensional text encoding $\Phi_\mathbf{T}\left(\mathbf{t}_c\right)$. 
Here $\mathbf{t}_c$ results from adding the class-specific word embedding $\mathbf{e}_c$ to a pre-defined prompt $\mathrm{p} $, i.e., $\mathbf{w}_c=\Phi_\mathbf{T}\left(\mathbf{t}_c\right)$ with $\mathbf{t}_c=\left\{\mathbf{p}, \mathbf{e}_c\right\}$. The prompt $\mathbf{p}$ is manually crafted to capture the semantic meaning of the downstream task, e.g., $\mathbf{t}_c=$ "A photo of a \{class\}". 
Given the image embedding $\mathbf{z}=\Phi_\mathbf{V}(\mathbf{x})$, the probability of the image $\mathbf{x}$ being classified as $y \in\{1, \ldots, C\}:=[C]$ is thus defined as 
\begin{equation} \label{clip-classifier}
\begin{split}
p\left(y \mid \mathbf{x}\right)=\frac{\exp \left( \mathbf{z}^\top \mathbf{w}_y / \tau\right)}{\sum_{i=1}^C \exp \left( \mathbf{z}^\top \mathbf{w}_i / \tau\right)}
\end{split}
\end{equation}
where $\tau$ is a temperature parameter.


\noindent \textbf{Federated Learning (FL).} Consider a federated learning scenario involving $K$ clients and a central server, and each client $k$ holds a local dataset $D^k=\left\{( \mathbf{x}_i^k, y_i^k)\right\}_{i=1}^{N^k}, k=1, \cdots, K$ containing $N^k$ samples. Let $D=\bigcup_{k=1}^K D^k$ represent the total datasets where each dataset is derived from a distinct data distribution $\mathcal{D}_k$.
Generally, federated learning is defined as an optimization problem~\cite{fedavg,fed-servey-open-problem-kairouz2021advances} for maximizing a global objective function $\mathbb{F}(\theta)$, which is a mixture of local objective functions $\mathbb{F}^k(\theta, D^k)$, namely $\mathbb{F}(\boldsymbol{\theta})=\sum_{k=1}^K \mathbb{F}_k\left(\boldsymbol{\theta}, D_k\right)$, where $\theta$ is the parameter vector of the global model.

\noindent \textbf{Notations.} In this paper, vectors are denoted by bold lowercase letters, and matrices by bold uppercase letters.
In the context of federated dataset, a superscript $k$ denotes data belonging to client $k$, a subscript $c$ denotes class‑$c$ samples.
For example, the data belonging to $c$-th class in total dataset is denoted as $D_c$, with its cardinality $N_c$. Similarly, the data of the $k$-th client belonging to class $c$ is denoted as $D^k_c$, and its cardinality as $N^k_c$.
The visual representation of $i$-th data $\mathbf{x}^k_{c,i}$ from $D^k_c$ can be computed as $\mathbf{z}_{c,i}^k = \Phi_\mathbf{V}(\mathbf{x}^k_{c,i})$.
\section{Methodology} \label{method}
\subsection{Overview}
In this section, we present the design of our \M~framework, as illustrated in Figure~\ref{framework-fig-1}, consisting of a visual pipeline, a textual pipeline, and an adaptive multimodal fusion module.
To \textbf{fully exploit the rich modal information in pre-trained VLMs} and \textbf{strike a balance between global generalization and local personalization}, \M~utilizes prompt distribution learning to extract personalized visual representations, enhances the robustness of LLM-based text augmentation through global alignment of importance scores, and integrates these components with an adaptive multimodal fusion mechanism.
To reduce communication costs and address data heterogeneity, without loss of generality, each client adopts a pre-trained CLIP model as the backbone model and has access to an LLM with consistent versions and synchronized parameters.
%
%
%
\M~is a training-free, one-shot federated framework that completes VLM adaptation in a single round without relying on gradient‑based model optimization, thereby bolstering both flexibility and implementation simplicity.
%

\subsection{Collaborative Prompt Distribution Learning}\label{Collaborative-Prompt-Distribution-Learning}

%
To capture the diverse visual variations, our method seeks to model the distribution of class prototypes in feature space. 
Prior research~\cite{GDA-CLIP-wang2024hard,prompt-distribution-learning-ProDA,Frolic-zhu2024enhancing-ProDA} have shown that Gaussian distributions effectively model the distribution of CLIP features, resulting in significant performance improvements.
%
%
Motivated by these observations, we assume $\mathcal{N} (\mathbf{w}_{1:C}, \mathbf{\Sigma})$ with identical covariance is the underlying class-specific prompt distribution over global and each client's local visual representations, where $\mathbf{w}_{c} \in \mathbb{R}^d$ presents the mean of the embedding distribution for the $c$-th class, $\mathbf{\Sigma}\in \mathbb{R}^{d\times d}$ denotes the shared covariance, and the density $\mathcal{N}\left(\mathbf{z} ; \mathbf{w}_c, \mathbf{\Sigma}\right)$ is
\begin{equation*} \label{gaussian}
\begin{split}
\frac{1}{\sqrt{(2 \pi)^d|\mathbf{\Sigma}|}} \exp \left\{-\frac{1}{2}\left(\mathbf{z}-\mathbf{w}_c\right)^{\top} \mathbf{\Sigma}^{-1}\left(\mathbf{z}-\mathbf{w}_c\right)\right\}.
\end{split}
\end{equation*}
For global dataset, the prompt distribution over visual representation $p(\mathbf{z} \mid y=c)$ approximated by $\mathcal{N} (\mathbf{w}_{c}, \mathbf{\Sigma})$.
%
%
To obtain personalized representations for each client, the visual feature embeddings $p(\mathbf{z}\mid y=c, D^k)$ for $c\in[C]$ are modeled as $\mathcal{N} (\mathbf{w}_{1:C}^k, \mathbf{\Sigma}^k)$, adopting the same distribution form as the global one.

Our goal is to extract personalized visual representations from the global features that are better adapted to the client's local objectives.
%
%
We first derive the parameters of the global prompt distribution $\theta=(\mathbf{w}_{1:C},\mathbf{\Sigma})$ using the Bayes' formula.
Given a prior distribution $\pi(\theta)$ on the parameters of the Gaussian distribution, the prompt prototypes of the global visual representation are derived from the posterior probability $q(\theta)=\pi(\theta \mid D)$. By Bayes' formula, 
\begin{equation} \label{global-bayes}
\begin{split}
    &\pi(\theta \mid D) \propto L(D \mid \theta) \pi(\theta),\\
\end{split}
\end{equation}
where $L(D \mid \theta):= \prod_{c \in [C]}\prod_{i=1}^{N_c} \mathcal{N}(\mathbf{z}_{c,i};\mathbf{w}_c, \mathbf{\Sigma})$ presents the global likelihood function given the parameter $\theta$.
Then, hierarchically, the personalized presentation extraction problem aims to deduce the local posterior distribution, with the global prompt distribution serving as the informative prior.
Denoting the local parameter as $\theta^k = (\mathbf{w}_{1:C}^k, \mathbf{\Sigma}^k)$, Bayesian inference suggests that the posterior
$$
q(\theta^k) \propto L(D^k \mid \theta^k)\pi(\theta^k) \propto L(D^k \mid \theta^k)L(D \mid \theta) \pi(\theta),
$$ 
where we plug in prior $\pi(\theta^k)=q(\theta)$, and $L(D^k \mid \theta^k)$ presents the local likelihood.
%

Hence, we adopt the power prior to ensure that the posterior is not overly impacted by global information, which introduces a scalar prior parameter $\alpha \in [0,1]$ that weights the prior distribution relative to the global likelihood.
The posterior distribution of $\theta^k$ can be written as 
\begin{equation} \label{local-bayes}
\begin{split}
q(\theta^k) \propto p(D^k \mid \theta^k)[L(D \mid \theta)]^{\alpha} \pi(\theta).
\end{split}
\end{equation}

The above personalized prompt distribution learning problem is equivalent to solving the local visual representation $q(\theta^k)$.
%
%
To ensure computational efficiency and facilitate one-shot adaptation, we design a conjugate prior to construct the hierarchical Bayesian framework with an explicit posterior formulation, as presented in the following lemma.

\begin{lemma} \label{lemma-1}
    Assume that the mean of each prompt prototype $\mathbf{w}_c$ is independent given shared covariance $\mathbf{\Sigma}$, the hierarchical Bayesian model characterized in~\eqref{global-bayes} and~\eqref{local-bayes} exists a conjugate prior $\pi(\theta)$ over parameter $(\mathbf{w}_{1:C},\mathbf{\Sigma})$: 
    \begin{align*}
    \mathbf{\Sigma} \sim \mathcal{I W}(\mathbf{\Sigma} ; \mathbf{S}_0, \nu_0), \quad \mathbf{w}_c| \mathbf{\Sigma} \sim \mathcal{N}(\mathbf{z} ; \mathbf{m}_{0,c}, \frac{1}{\kappa_{0,c}} \mathbf{\Sigma}), 
    \end{align*}
    where $\ c\in [C]$ and $\mathcal{I W}(\cdot)$ denote the Inverse-Wishart distribution. Specifically, denoting the sample number count for class $c$ as $N_c$ and the embedding for the $i$-th sample in class $c$ as $\mathbf{z}_{c,i}$, the posterior distribution can be formulated as $\mathcal{N}\left(\mathbf{z} ; \mathbf{w}_{1:C}^*, \mathbf{\Sigma}^*\right)$, and for $c\in[C]$, 
    \begin{align*}
    \mathbf{\Sigma^*} \sim \mathcal{I W}(\mathbf{\Sigma} ; \mathbf{S}_q, \nu_q), \quad \mathbf{w}^*_c| \mathbf{\Sigma^*} \sim \mathcal{N}(\mathbf{z} ; \mathbf{m}_{q,c}, \frac{1}{\kappa_{q,c}} \mathbf{\Sigma^*}). 
    \end{align*}
    The parameters are specified as
    \begin{flalign}\small\label{posterior-formula-main}
    &\kappa_{q,c} = \kappa_{0,c}+ N_c, \qquad \mathbf{m}_{q,c} = \frac{\kappa_{0,c} \mathbf{m}_{0,c}+ N_c \bar{\mathbf{z}}_c}{N_c+\kappa_{0,c}} \notag\\
    &\nu_q = \nu_0 + \sum_{c \in [C]} N_c\\
    &\mathbf{S}_q = \mathbf{S}_0+\sum_{c \in [C]}\mathbf{S}_c  + \sum_{c \in [C]}\big(\kappa_{0.c} \mathbf{m}_{0,c}\mathbf{m}_{0,c}^\top - \kappa_{q,c} \mathbf{m}_{q,c}\mathbf{m}_{q,c}\big), \notag
    \end{flalign}
    where $\mathbf{S}_c:=\sum_{i=1}^{N_c} \mathbf{z}_{c,i}\mathbf{z}_{c,i}^\top$ and $\bar{\mathbf{z}}_c:= \frac{1}{N_c}\sum_{i=1}^{N_c}\bar{\mathbf{z}}_{c,i}$.
\end{lemma}
For the above hierarchical Bayesian model, an uninformative prior has the form $\mathbf{S}_0=s_0 \mathbf{I}, \mathbf{m}_{c,0}=\mathbf{0}, \nu_0=0$ and $s_0 \approx \kappa_{c,0} \approx0$ to some small positive number.
We now provide a detailed description of each step in the visual pipeline:
%
\begin{description}[labelwidth=3em, labelsep=0.5em, leftmargin=!, align=parleft]
  \item[\textbf{Step 1}:] Clients transmit local statistics to the server, which computes the posterior under uninformative prior via Lemma~\ref{lemma-1}, as global class-specific prototypes characterized by $\mathbf{S}_g, \nu_g, \mathbf{m}_{g,c},\kappa_{g,c}$ for $c\in[C]$.
  \item[\textbf{Step 2}:] The server sends global prototypes to each client.
  \item[\textbf{Step 3}:] Each client $k$ derives personalized, class-specific prototypes, parameterized by $\mathbf{S}_l^k, \nu_l^k, \mathbf{m}_{l,c}^k, \kappa_{l,c}^k$, for $c\in [C]$, using the global prototypes distribution as the prior via Lemma~\ref{lemma-1}.
\end{description}
The proof of Lemma 1 and the detailed expressions for the global and personalized prompt prototype parameters are provided in \textbf{Appendix B.1} with the \textbf{computation and privacy analysis}.

%
%

%
%
Furthermore, we utilize the maximum a posteriori (MAP) estimation of the local prompt distribution for the hierarchical model presented above,
\begin{equation*} \label{local-posterior-distribution}
\begin{split}
\underset{\mathbf{m},\mathbf{\Sigma}}{\arg \max} \ f_{p(\mathbf{w}_{l,c}^k, \mathbf{\Sigma}^k)}( \mathbf{m},\mathbf{\Sigma} | D, D^k) = (\mathbf{m}_{l,c}^k, \frac{\mathbf{S}_l^k}{\nu_l^k + d +2}).
\end{split}
\end{equation*}
If $\widehat{\mathbf{\Sigma}}^k_l= \mathbf{S}_l^k / (\nu_l^k + d +2))$ is positive definite, \eqref{local-posterior-distribution}~indicates that $\mathcal{N}(\mathbf{z};\mathbf{m}_{l,c}^k, \widehat{\mathbf{\Sigma}}^k_l)$ is the most probable distribution within the Gaussian family.
%
%
By applying Gaussian Discriminant Analysis (GDA), the personalized classification probability $f^k_V(\mathbf{z})$ can be formulated as 
\begin{equation*} \small
\begin{split}
p(&y =c \mid \mathbf{z}, D^k) = \frac{p(\mathbf{z}\mid y=c,D^k)p(y=c\mid D^k )}{\sum_{i=1}^C p(\mathbf{z}\mid y=i,D^k)p(y=i\mid D^k )}\\
&= \frac{\exp \left((\mathbf{m}_{l,c}^k)^\top \mathbf{G}^k \mathbf{z}-\frac{1}{2} (\mathbf{m}_{l,c}^k)^\top \mathbf{G}^k \mathbf{m}_{l,c}^k+\log p_c^k\right)}{\sum_{i=1}^C \exp \left((\mathbf{m}_{l,i}^k)^\top \mathbf{G}^k \mathbf{z}-\frac{1}{2} (\mathbf{m}_{l,i}^k)^\top \mathbf{G}^k \mathbf{m}_{l,i}^k+\log p_i^k\right)},
\end{split}
\end{equation*}
where $\mathbf{G}^k:=\left(\widehat{\mathbf{\Sigma}}^k_l\right)^{-1}$, and $p_c^k = p(y=c \mid D^k):= 1/C$ is the class-wise prior probability for $c\in [C]$, which is set to be uniform during the prediction phase.

\subsection{Text Augmentation with Global Alignment}
The textual pipeline produces globally robust text prompts to complement personalized prompt distributions and mitigate data heterogeneity. 
Each client uses a local LLM to generate augmented descriptions, evaluates their reliability, and assigns weights. 
A global alignment procedure then aggregates these weighted descriptions into a robust, dataset-wide text augmentation.
To enhance the textual modality, we adopt the two-step prompt augmentation from~\cite{awt-zhu2024awt-nips2024}, using LLMs to generate dataset-aware class descriptions. LLMs produce probing questions from dataset-level summaries. These are then combined with class names to form tailored descriptions, ensuring both diversity and visual relevance. For each class $c$, the augmented text set is $\{\mathbf{t}_c^m\}_{m=1}^M$, supplemented by the manual prompt “A photo of a {class}” denoted as $\mathbf{t}_c^0$.

Subsequently, we propose a global text prompt alignment method within FL environments, designed to extract text augmentations that exhibit both generalization and robustness at the global level.
%
To determine the importance of text description for the $j$-th class, each client calculates the classification probability with the text input $\mathbf{t}$:
\begin{equation}
\begin{split}
p^k_c(\mathbf{t}) =\frac{\exp \left(\frac{1}{N_c^k}\sum_{i=1}^{N^c_j} \Phi_\mathbf{T}(\mathbf{t})^\top\mathbf{z}^k_{c,i}\right)}{\sum_{c'=1}^C \exp \left(\frac{1}{N_{c'}^k}\sum_{i=1}^{N_{c'}^k}   \Phi_\mathbf{T}(\mathbf{t})^\top\mathbf{z}^k_{c',i}\right)}.
\end{split}
\end{equation}
With augmented texts $\mathbf{t}_c^m$, $p^k_j(\mathbf{t}_{c}^m)$ represents the significance of the text prompt $\mathbf{t}_c^m$ for class $j$, as evaluated from the visual feature embeddings. 
For the robust text augmentation of $c$-th class, the most important property is its capability to distinguish the $c$-th class's image features from those of other classes.
Therefore, on the server side, we introduce a significance scoring criterion based on the manually defined prompts $\mathbf{t}_c^0$,
\begin{equation} \label{text-score}
\begin{split}
r(\mathbf{t}_c^m) = \frac{1}{K}\sum_{k=1}^K u^k(\mathbf{t}_c^0) \log\left( \frac{u^k(\mathbf{t}_c^m)}{u^k(\mathbf{t}_c^0)} \right),
\end{split}
\end{equation}
where $u^k(\mathbf{t}_c^m):=p^k_c(\mathbf{t}_c^m) -\max_{j \neq c} p^k_j(\mathbf{t}_c^m)$ indicates the confidence of the text input $\mathbf{t}_c^m$ in local classification tasks.
Since manually designed inputs $\mathbf{t}_c^0,\ c\in [C]$ are considered the most robust text inputs across various environments,  the score function~\eqref{text-score}, which is analogous to the KL divergence, assigns a higher importance score to text prompts with stronger robustness, approaching or exceeding that of $\mathbf{t}_c^0$, in federated downstream tasks.
We then weight the descriptions within the $c$-th class based on the importance scores as follows:
\begin{equation} \label{text-score-1}
\begin{split}
\mathbf{b}(\mathbf{t}_c^m) = \frac{\exp\left( r(\mathbf{t}_c^m) / \tau_t \right)}{\sum_{m=0}^M\exp\left( r(\mathbf{t}_c^m) / \tau_t \right)}, \quad m=0,\dots,M,
\end{split}
\end{equation}
where $\tau_t$ is the temperature parameter.
Here we set $\tau_t=0.5$ to assign a higher weight to robust text prompts.
During the classification phase, denoting $\mathbf{z}=\Phi_\mathbf{V}(\mathbf{x})$, the prediction probability of image $\mathbf{x}$ being classified as class 
$c$ can be computed as
\begin{equation} \label{text-classifier}
\begin{split}
f_T(\mathbf{z}) = \frac{\exp \left( \sum_{m=0}^M \mathbf{b}(\mathbf{t}_c^m)\mathbf{z}^\top \Phi_\mathbf{T}(\mathbf{t}_c^m) \right)}{\sum_{j=1}^C \exp \left( \sum_{m=0}^M \mathbf{b}(\mathbf{t}_c^m) \mathbf{z}^\top \Phi_\mathbf{T}(\mathbf{t}_j^m) \right)}.
\end{split}
\end{equation}

\subsection{Adaptive Multimodals Fusion}

As a general rule within FL, combining the global robust model with locally personalized models can further improve performance under data heterogeneity~\cite{ditto-li2021ditto,fedprox-li2020federated,pfedprompt-guo2023pfedprompt,ref-10,Frolic-zhu2024enhancing-ProDA}.
%
%
%
To further enhance the effectiveness of our method in data heterogeneous environments, we propose a sample-wise ensembling technique that adaptively calibrates inter-modal weights to fuse the personalized visual representations and robust language modalities.
The key in our prediction fusion lies in introducing a sample-wise mixing coefficient $\eta(\mathbf{z})$ to balance the contributions of both modalities, namely
$$
f_M^k(\mathbf{z}) = \eta(\mathbf{z})f^{k}_{V}(\mathbf{z}) + (1-\eta(\mathbf{z}))f_T(\mathbf{z}).
$$
The following theorem provides the theoretical motivation for this module’s design.
\begin{theorem} \label{theorem-1}
    Let $f(\mathbf{z}):=\eta(\mathbf{z})f_1(\mathbf{z}) + (1-\eta(\mathbf{z}))f_2(\mathbf{z})$, where $f_1,f_2 \in \mathcal{H}$ are two prediction functions, and $\mathcal{H}: \mathcal{X} \rightarrow\{-1,+1\}$ is a hypothesis set. Denoting the empirical predictive errors on $\mathcal{D}_{\text {train }}=\left\{\mathbf{z}_i, y_i\right\}_{i=1}^N$ as $\widehat{\mathcal{R}}\left(f_i\right),i=1,2$, and the VC dimension of $\mathcal{H}$ as $d_{VC}(\mathcal{H})$, then with probability at least $1-\delta$ over the samples,
    \begin{align*}
        \mathcal{R}(f) &\leq B \sqrt{\frac{d_{VC}(\mathcal{H})+\log 1/\delta}{N}} +\sum_{i=1,2}\widehat{\mathcal{R}}(f_i)\\
        &\quad+ \operatorname{Cov}\left(\eta(\mathbf{z}), \ell_1(\mathbf{z})-\ell_2(\mathbf{z})\right),
    \end{align*}
    where $\mathcal{R}(f)=\mathbb{E}_{(\mathbf{z},y)\sim \mathcal{D}}[\ell(f(\mathbf{z}),y)]$, $\ell$ is the cross-entropy loss, $\operatorname{Cov}$ denotes the covariance and $B$ is a constant.
\end{theorem}
The proof is detailed in Appendix B.2. According to the above expression, minimizing the generalization error requires $\eta$ to be proportional to $\ell_2-\ell_1$, while $\ell:= -\log p_{true}$.

%
%
\citet{well-calibrated-kumar2022calibrated} have shown that a well-calibrated classifier’s confidence serves as a reasonable surrogate for its true accuracy $p_{true}$.
Formally, the average confidence over the dataset $\{x_i\}_{i=1}^N$ scaled by a temperature $\tau >0$ is given by the average of the model’s probability for its prediction:
$$\operatorname{conf}(f, \tau)=\frac{1}{N} \sum_{i=1}^N \max _j \operatorname{softmax}\left(f\left(x_i\right) / \tau\right)_j.$$
To approximate a well-calibrated classifier, the average confidence of model $f$ should reflect the predicted accuracy, i.e., $\operatorname{conf}(f, \tau) \approx \operatorname{Acc}(f)$.
This can be implemented by binary search of $\tau$, which works since the confidence increases when $\tau$ decreases.
%
%
Aiming to ensure that the fused classifier minimizes the generalization error, the weight $\eta(\mathbf{z})$ can be designed as
$$
\eta(\mathbf{z}) = \frac{1}{1+e^{-L(\mathbf{z})}}, \ L(\mathbf{z}):=\log\left(\frac{\max_j\operatorname{softmax}(f_V^k(\mathbf{z}))_j}{\max_j\operatorname{softmax}(f_T(\mathbf{z}))_j}\right).
$$
%
%
Thus, the fused classifier naturally favors the modality with higher predictive accuracy for given regions, thereby enhancing overall performance.

\section{Experiment}
%
\begin{table*}[t]
\centering
  \caption{Few-shot Performance on CLIP Datasets over 10 Clients.}
  \vspace{-5pt}
    \begin{tabular}{l|c|c|ccccc}
    \toprule
    Method & Training-free & One-shot & OxfordPets & Flowers102 & Food101 & Caltech101 & DTD \\
    \midrule
    CoOp  & \XSolidBrush & \XSolidBrush & 89.18 & 69.03 & 82.54 & 90.62 & 63.97 \\
    PromptFolio & \XSolidBrush & \XSolidBrush & 92.08 & 74.61 & 86.50 & 93.59 & 65.04 \\
    DP-PFL & \XSolidBrush & \XSolidBrush & \textcolor[rgb]{ .184,  .459,  .71}{96.91} & 85.75 & 86.08 & \textcolor[rgb]{ .184,  .459,  .71}{96.76} & \textcolor[rgb]{ .184,  .459,  .71}{86.23} \\
    PromptFL & \XSolidBrush & \XSolidBrush & 90.79 & 92.26 & 88.17 & 87.90 & 50.46 \\
    pFedPromp & \XSolidBrush & \XSolidBrush & 91.84 & \textcolor[rgb]{ .184,  .459,  .71}{96.46} & \textcolor[rgb]{ .184,  .459,  .71}{92.26} & 96.54 & 77.14 \\
    \cline{1-8}
    Zero-Shot CLIP & \Checkmark & \Checkmark & 85.77 & 66.14 & 77.31 & 86.29 & 42.32 \\
    CLIP-GDA & \Checkmark & \Checkmark & 88.81 & 91.23 & 79.05 & 92.55 & 60.64 \\
    FedLPA+PromptFL & \XSolidBrush & \Checkmark & 83.42 & 78.60 & 74.74 & 88.69 & 52.75 \\
    FENS+PromptFL & \XSolidBrush & \Checkmark & 90.51 & 81.19 & 80.80 & 84.37 & 68.43 \\
    FedBEns+PromptFL & \XSolidBrush & \Checkmark & 79.84 & 86.27 & 78.45 & 86.58 & 65.82 \\
    \rowcolor[rgb]{ .851,  .851,  .851} TOFA (Ours) & \Checkmark & \Checkmark & \textbf{91.23} & \textbf{95.78} & \textbf{85.49} & \textbf{94.58} & \textbf{71.68} \\
    \bottomrule
    \end{tabular}%
    \begin{tablenotes}[label={}]
        \item \quad \ * \textcolor[rgb]{ .184,  .459,  .71}{Blue} denotes the highest results of multi-round methods. \textbf{Bold} denotes the highest results of one-shot methods.
    \end{tablenotes}
  \label{tab:overall-performance}%
  \vspace{-10pt}
\end{table*}%


\subsection{Setup}
Due to space limitations, the detailed information in this section is provided in \textbf{Appendix C}.

\noindent\textbf{Datasets.} We assess the effectiveness of the proposed \M~across nine publicly available benchmark datasets under various federated configurations to simulate different types of data heterogeneity:
%
(1) Five representative visual classification datasets commonly employed to test few-shot performance in the CLIP benchmark~\cite{CLIP}: \textbf{OxfordPets}, \textbf{Flowers102}, \textbf{DTD}, \textbf{Caltech101}, and \textbf{Food101}, hereafter referred to collectively as the CLIP datasets.
%
%
(2) Two standard image benchmarks, \textbf{CIFAR‑10} and \textbf{CIFAR‑100}~\cite{cifar-gong2012geodesic}.
%
%
(3) Two multi-domain datasets with feature shift~\cite{fedBN-li2021fedbn}: \textbf{DomainNet}~\cite{domainnet-peng2019moment} (six domains), and \textbf{Office‑Caltech10}~\cite{cifar-gong2012geodesic} (four domains).
%

%
\noindent \textbf{Baselines.} Since \textit{no previous work was found that investigates training-free and one-shot distributed adaptation method for VLMs} within a FL framework, we compare the performance of \M~with four categories of baselines: 
(1) Four existing prompt learning federated learning methods:  \textbf{PromptFolio}~\cite{portfolio-fed-theory-pan2024federated}, \textbf{DP-PFL}~\cite{DP-FPL-tran2025privacypreserving}, \textbf{PromptFL}~\cite{promptfl-guo2023promptfl}; \textbf{pFedPrompt}~\cite{pfedprompt-guo2023pfedprompt}. 
(2) Three local adapting methods: \textbf{Zero-shot CLIP}~\cite{CLIP} with hand-crafted text prompt templates; \textbf{CLIP-GDA}~\cite{GDA-CLIP-wang2024hard}; \textbf{CoOp}~\cite{COOP-zhou2022learning}.
(3) Three adapted methods derived from advanced one-shot techniques that operate solely with client-side training resources, combined with the backbone prompt learning method~\cite{promptfl-guo2023promptfl} in FL: \textbf{FedLPA}~\cite{oneshot-fedlpa-liu2024fedlpa}, \textbf{FENS}~\cite{FENS-allouah2024revisiting}, \textbf{FedBEns}~\cite{oneshot-FedBEns-talpini2025fedbensoneshotfederatedlearning}.
(4) \textbf{FedAvg}~\cite{fedavg} is included as a  traditional baseline in experiments on image datasets.

\noindent \textbf{Implementation details.} (1) CLIP datasets. Each dataset in CLIP datasets is partitioned into $N = 10$ clients, defaulting if not explicitly specified, each with a disjoint set of classes evenly and randomly assigned to the clients.
(2) CIFAR10 and CIFAR100. We split $N = 100$ clients resulting from $Dir(\beta = 0.3)$ partition.
(3) DomainNet and Office-Caltech10. Each client in the federated system is assigned data from a single unique domain, Consistent with prior work~\cite{fedBN-li2021fedbn,fedpgp-harmon-pmlr-v235-cui24c}. 
We present the results using ViT-B16~\cite{vit-b16-dosovitskiy2020image} backbones.
%
For other hyperparameters, such as learning rate and local epochs in the aforementioned baselines, we adhere to the original configurations from these studies.
Given that the results of our model are deterministic due to its training-free nature, we do not present results with statistical variations, which is typical in zero-shot or training-free studies~\cite{Frolic-zhu2024enhancing-ProDA,awt-zhu2024awt-nips2024,GDA-CLIP-wang2024hard}.
%

\subsection{Overall Comparison}

\noindent \textbf{Model evaluation on label shifts.} We began by assessing performance of \M~against baselines on datasets with label shift.
We conducted few‑shot experiments on the CLIP datasets and standard training experiments on the CIFAR datasets.
In Table~\ref{tab:overall-performance}, we present the numerical results of the training-required/training-free and multi-round/single-round baselines under the 16-shot setting.
Results show that \M~consistently exceeds the performance of one-shot baselines on all five datasets, even exceeds many muti-round prompt learning methods, owing to full exploration of modal interaction information from pre-trained VLMs.
For example, our method consistently outperforms CoOp, PromptFolio, and PromptFL across five vision datasets, except for Food101.
Notably, on the DTD dataset, where training-free methods struggle to perform well, \M~still outperforms all one-shot baselines and shows only a slight performance gap compared to multi-round methods like PromptFolio and PromptFL.
Demonstrating \M’s remarkable ability to adapt to resource-constrained scenarios in few-shot settings.
Table~\ref{tab:Cifar-result} presents the comparison results on CIFAR datasets, which are partitioned under Dirichlet setting $\beta=0.3$ over 100 clients. 
%
%
This result further corroborates the efficacy of our method in handling extreme data heterogeneity.
It also empirically demonstrates the scalability of \M~with respect to the number of clients.

\begin{table}[h]
  \centering
  \vspace{-5pt}
  \caption{Results on CIFAR10 \& CIFAR100}
  \vspace{-5pt}
    \begin{tabular}{l|cc}
    \toprule
    Method & Cifair10 & Cifair100 \\
    \midrule
    FedAvg & 75.10 & 42.52 \\
    Zero-Shot CLIP & 87.71 & 64.92 \\
    CoOp  & 93.11 & 74.83 \\
    PromptFL & 92.30 & 73.67 \\
    \midrule
    TOFA (Ours) & 93.18 & 76.63 \\
    \bottomrule
    \end{tabular}%
    \vspace{-5pt}
  \label{tab:Cifar-result}%
\end{table}%

\noindent \textbf{Model evaluation on feature shifts.} To assess the performance of our method in scenarios more closely resembling real-world FL applications, we examine \M~on real-world data with feature shift~\cite{fedBN-li2021fedbn} using DomainNet and Office-Caltech10, where each client is assigned with single domain dataset, resulting in 6 clients for DomainNet and 4 clients for Office-Caltech10.
\begin{table*}[h]
  \centering
  \caption{Experimental Results on Office-Caltech10 and DomainNet Datasets with Feature Shift.}
  \vspace{-5pt}
    \begin{tabular}{l|ccccccc|ccccc}
    \toprule
    \multicolumn{1}{l}{Datasets} & \multicolumn{7}{c}{DomainNet}                         & \multicolumn{5}{c}{Office-Caltech10} \\
\cmidrule{2-13}    \multicolumn{1}{l}{Domains} & C     & I     & P     & Q     & R     & S     & \multicolumn{1}{c}{Avg.} & A     & C     & D     & W     & Avg. \\
    \midrule
    CoOp  & 98.32 & 83.01 & 98.18 & 82.37 & 98.21 & 97.70 & 92.97 & 96.38 & 97.24 & 100   & 98.31 & 97.98 \\
    PromptFolio & 98.38 & 83.07 & 98.24 & 82.43 & 98.27 & 97.84 & 93.04 & 95.64 & 96.50 & 99.26 & 97.57 & 97.24 \\
    DP-PFL & \textcolor[rgb]{ .184,  .459,  .71}{98.93} & \textcolor[rgb]{ .184,  .459,  .71}{84.52} & \textcolor[rgb]{ .184,  .459,  .71}{98.89} & \textcolor[rgb]{ .184,  .459,  .71}{87.87} & \textcolor[rgb]{ .184,  .459,  .71}{98.64} & 98.02 & \textcolor[rgb]{ .184,  .459,  .71}{94.48} & \textcolor[rgb]{ .184,  .459,  .71}{97.92} & 97.68 & \textcolor[rgb]{ .184,  .459,  .71}{100} & \textcolor[rgb]{ .184,  .459,  .71}{100} & \textcolor[rgb]{ .184,  .459,  .71}{98.90} \\
    PromptFL & 98.23 & 79.91 & 97.89 & 66.52 & 96.83 & 97.31 & 89.45 & 96.41 & 96.39 & 96.90 & 100   & 97.43 \\
    pFedPromp & 98.14 & 82.43 & 98.26 & 86.52 & 96.98 & \textcolor[rgb]{ .184,  .459,  .71}{98.42} & 93.46 & 97.12 & \textcolor[rgb]{ .184,  .459,  .71}{98.18} & 96.85 & 100   & 98.04 \\
    \midrule
    Zero-Shot CLIP & 72.32 & 47.15 & 53.63 & 31.30 & 48.40 & 50.18 & 50.50 & 19.30 & 18.20 & 21.90 & 18.60 & 19.50 \\
    CLIP-GDA & 71.72 & 60.75 & 64.39 & 67.70 & 66.48 & 69.19 & 66.71 & 96.08 & \textbf{98.20} & 98.30 & 100   & 98.15 \\
    Oneshot+PromptFL & 85.42 & 75.68 & 86.97 & 73.96 & 84.74 & 89.61 & 82.73 & 94.93 & 96.21 & 96.89 & 99.20 & 96.81 \\
    \rowcolor[rgb]{ .851,  .851,  .851} TOFA (Ours) & \textbf{98.86} & \textbf{82.69} & \textbf{97.45} & \textbf{83.37} & \textbf{98.12} & \textbf{97.83} & \textbf{93.05} & \textbf{96.94} & 97.81 & \textbf{100} & \textbf{100} & \textbf{98.69} \\
    \bottomrule
    \end{tabular}%
    \begin{tablenotes}[label={}]
         \item * \textcolor[rgb]{ .184,  .459,  .71}{Blue} denotes the highest results of multi-round training methods. \textbf{Bold} denotes the highest results of one-shot methods.
    \end{tablenotes}
    \vspace{-5pt}
  \label{tab:domain-comparison}%
\end{table*}%

\begin{figure*}[t]
\centering
\includegraphics[width=0.95\linewidth]{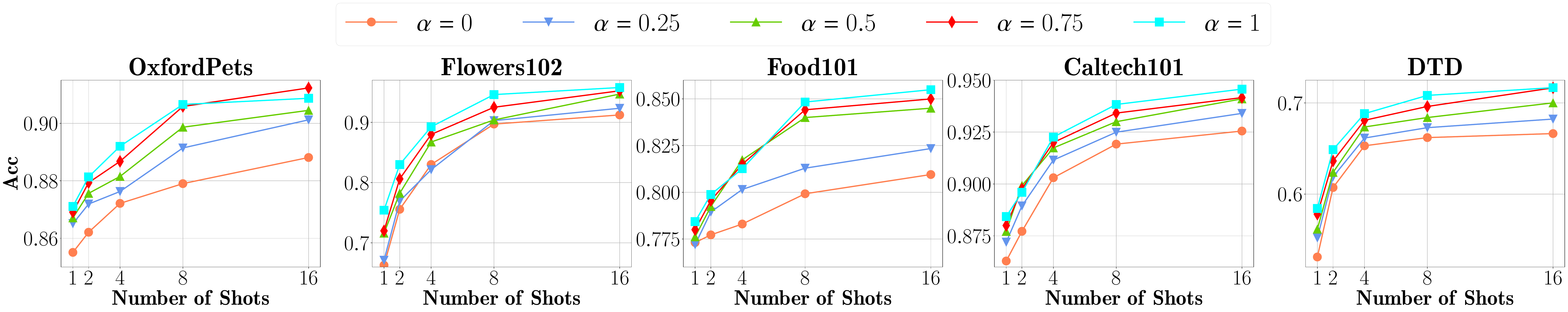} 
\vspace{-5pt}
\caption{Comparisons on CLIP datasets across varying shot numbers and parameter $\alpha$ in \M~over 10 clients.}
\vspace{-5pt}
\label{shot-numbers-fig-1}
\end{figure*}
In Table~\ref{tab:domain-comparison}, we report the performance of our method against other baselines on these two domain datasets.
We present the maximum accuracy of these methods combining one-shot techniques with prompt-based FL (Oneshot+PromptFL).
The experimental results show that training-free methods, such as Zero-Shot CLIP and CLIP-GDA, struggle to benefit the clients, particularly on the DomainNet dataset.
However, our method achieved the highest average accuracies $93.05\%$ and $98.69\%$ on Office-Caltech10 and DomainNet, respectively. 
This result surpasses most methods requiring multiple rounds of training and achieves a performance within $2\%$ of the optimal prompt-based FL baseline. %
This validates the effectiveness and robustness of \M~in scenarios closer to real-world federated settings.
\subsection{Ablation Experiments}
\noindent \textbf{Impact of number of $\alpha$.} In \M, $\alpha \in [0,1]$ is the coefficient for adjusting the contribution of global information in local distribution posterior inference. 
Figure~\ref{shot-numbers-fig-1} displays the results for $\alpha$ values ranging from $0$ to $1$ across the CLIP datasets.
The trends reveal that although the optimal $\alpha$ value varies across datasets, assigning a higher weight to global information (e.g. $\alpha\geq0.75$) can achieve near-optimal performance.
Based on this observation, we adopt $\alpha=1$ for experiments in this section.

\noindent \textbf{Impact of number of shot.} We also investigate the impact of shots in the few-shot learning for \M. 
Figure~\ref{shot-numbers-fig-1} also presents the impact of shots in the few-shot learning for~\M across the CLIP datasets. The number of shots varies from $[1, 2, 4, 8,16]$.
The result shows an explicit improvement in test accuracy with an increase in the number of shots.
Based on the accuracy shown in the figure, our method achieves stable results starting from the 8-shot classification.

\begin{figure}[t]
\centering
\includegraphics[width=0.95\linewidth]{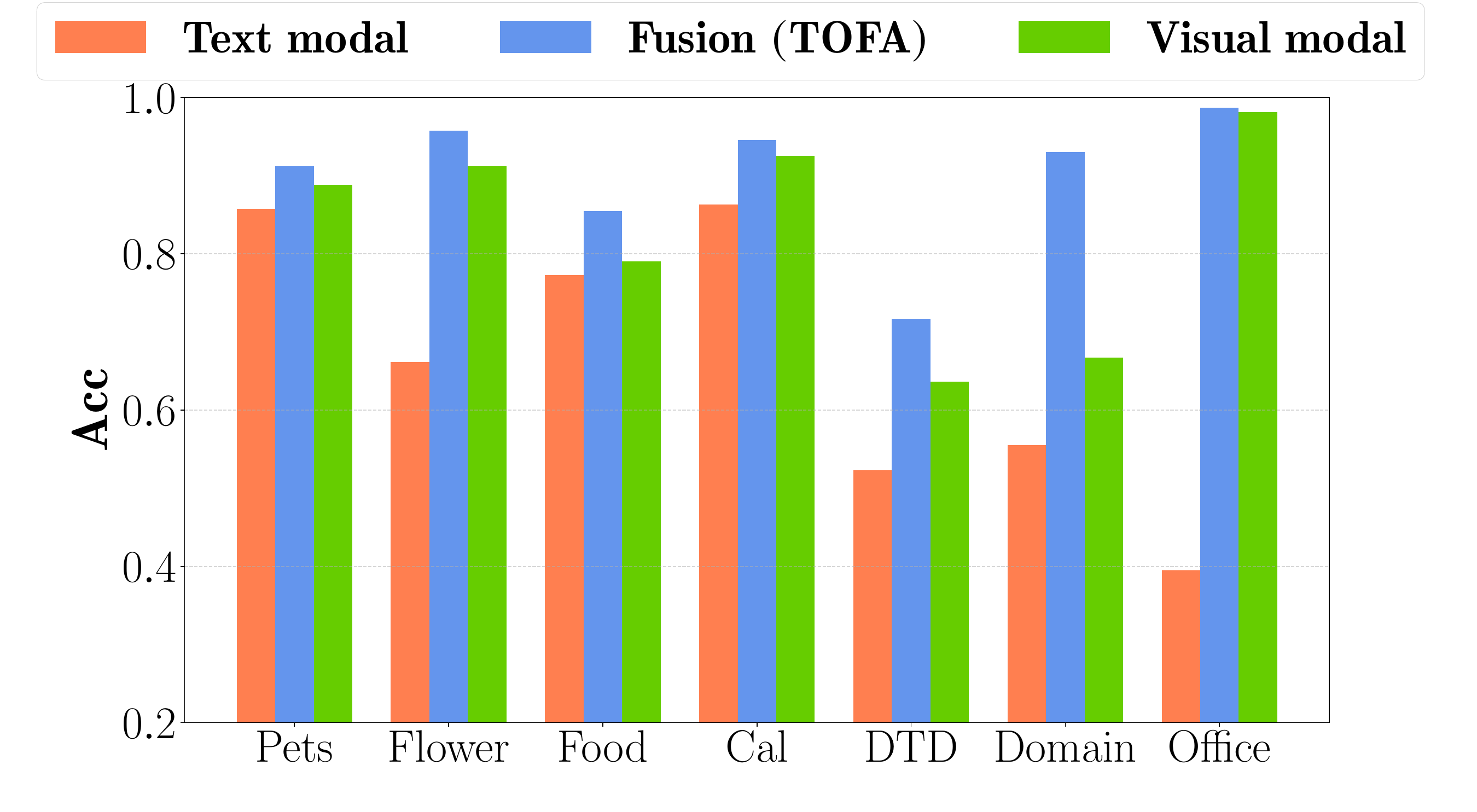} 
\vspace{-5pt}
\caption{Results on Inter-Modality Ablation Experiments.}
\vspace{-5pt}
\label{model-abultion-fig}
\end{figure}

\noindent \textbf{Inter-modality ablation experiments.} We conducted an inter-modality ablation study to analyze the impact of different modalities on the overall performance of TOFA, as presented in Figure~\ref{model-abultion-fig}. 
The results indicate that the accuracy of both the visual and textual modalities before fusion is lower than that of the fused TOFA model. 
This demonstrates that combining personalized visual information with robust textual prompts effectively prevents overfitting in the fused model, thereby improving accuracy.
It further validates the importance of multimodal information fusion within downstream tasks of VLMs, which, compared to single-modal approaches, enables models to capture more precise and generalizable representations.


\section{Conclusion}
We propose~\M, a novel Training-free One-shot Federated Adaptation framework for VLMs in federated learning. To our knowledge, it is the first approach to realize training-free VLM adaptation with a single communication round. \M~leverages both visual and textual pipelines to extract task-specific, generalizable multimodal representations. In the visual pipeline, a hierarchical Bayesian model learns personalized, class-specific prototype distributions over visual features. In the textual pipeline, TOFA evaluates and globally aligns generated descriptions to ensure robustness. An adaptive weight calibration mechanism then fuses predictions from both modalities, trading off personalization and robustness to mitigate data heterogeneity. The method requires no additional model training on either clients or the server. Extensive experiments on nine datasets under diverse federated settings demonstrate the effectiveness of~\M.
\clearpage

\section*{Acknowledgments}
This work was supported in part by National Key R\&D Program of China (2024YFB3908400, 2024YFB3908403), and the National Natural Science Foundation of China (No.62172362).

\bibliography{aaai2026}



\clearpage

\appendix

\section{Related Work} \label{appendix-related-work}
\noindent \textbf{Vision-Language Models.}  In recent years, vision-language models (VLMs) have emerged as a foundational paradigm for unifying visual and textual understanding by pre‑training on massive, web‑scale image–text pairs. 
By jointly encoding images and their associated captions via separate vision and language encoders and optimizing a contrastive alignment objective, methods like CLIP~\cite{CLIP} and ALIGN~\cite{ALIGN-pmlr-v139-jia21b} learn an aligned multimodal embedding space in which semantically related inputs lie close together. 
This unified representation endows VLMs with remarkable transfer capabilities: they support zero‑shot and few‑shot classification, in‑context adaptation, and open‑world recognition of novel concepts without task‑specific fine‑tuning~\cite{CLIP,SigLIP-Zhai_2023_ICCV,vlm-relatedwork-1-pmlr-v139-kim21k,vlm-relatedwork-2-Su2020VL-BERT,vlm-relatedwork-3-pmlr-v139-jia21b,prompt-distribution-learning-ProDA,GDA-CLIP-wang2024hard,awt-zhu2024awt-nips2024,COOP-zhou2022learning}.
Beyond image classification, these models have been extended to open‑set object detection~\cite{object-detection-1-feng2022promptdet,object-detection-2-du2022learning,object-detection-3-gupta2022ow} and segmentation~\cite{segment-1-ding2022decoupling,segment-2-xu2022simple}, video action recognition~\cite{video-action-1-wang2021actionclip,video-action-2-weng2023open,video-action-3-weng2023open}, and even conditional image generation~\cite{image-generation-1-ramesh2022hierarchical,image-generation-2-rombach2022high,fengcontrollable}, demonstrating the flexibility of natural language prompts to specify a diverse array of vision tasks.


\noindent \textbf{VLMs Adaptation in Federated Learning.} In federated learning (FL), VLM adaptation predominantly relies on fine‑tuning and prompt learning techniques. 
In fine‑tuning paradigm, clients collaboratively update the full or partial weights of pre‑trained VLMs through multiple communication rounds, which can yield high task‑specific accuracy but incurs substantial computation and communication overhead~\cite{fed-fine-tuning-1-nguyen2024flora,fed-fine-tuning-2-10041745,fed-fine-tuning-3-10776777,fedllm}. 
Prompt-based federated learning offers a lightweight adaptation by learning and aggregating small prompt vectors instead of full model parameters
For example, to accelerate the global aggregation and handle situations with insufficient user data, FedPrompt~\cite{fedprompt-iccsp-10095356} and PromptFL~\cite{promptfl-guo2023promptfl} have introduced prompt learning into Federated Learning.
To tackle the statistical heterogeneity often found in client data, pFedPrompt~\cite{pfedprompt-guo2023pfedprompt} introduces a non-parametric approach, providing each client with a personalized attention module.
Recently, PromptFolio~\cite{portfolio-fed-theory-pan2024federated} introduces a global-local prompt portfolio that integrates global task-relevant features with personalized local features.
%
%
DP‐FPL~\cite{DP-FPL-tran2025privacypreserving} proposes a privacy-preserving federated prompt learning model that factorizes local prompts with a residual term to balance personalization and generalization.
While these approaches have achieved satisfactory results, they still require additional training processes and multi-round communication to learn the prompt vectors.
Contrary to these existing methods, this work introduces a training‑free one-shot adaptation approach that achieves model accuracy comparable to multi‑round prompt learning.

\noindent \textbf{One-Shot Federated Learning.}
One‑shot Federated Learning (FL) is an emerging research paradigm distinguished by its ability to achieve model adaptation with only a single round of communication.
The inaugural work~\cite{one-shot-begin-origin-guha2019oneshotfederatedlearning} ensembled locally trained models into a global predictor. 
Subsequently, various techniques were introduced to enhance the capabilities of one-shot federated training, e.g. data distillation~\cite{oneshot-distill-1-li2020practical,oneshot-distill-2-lin2020ensemble}, XOR‑based sample encoding~\cite{oneshot-xor-shin2020xor} and synthetic data generation on the server~\cite{oneshot-generate-1-zhang2022dense,oneshot-generate-3-diao2023towards,oneshot-generate-2-dai2024enhancing}
FedDISC~\cite{feddisc-Yang_Su_Li_Xue_2024} proposes to adapt Diffusion model for classification without client training, but requires labeled data and training resources on the global server, making it unsuitable for federated learning scenarios with limited server capabilities.
Recently, there has been increasing attention on efficient one-shot training methods that do not require server-side data or training resources~\cite{FENS-allouah2024revisiting,oneshot-fedlpa-liu2024fedlpa,oneshot-FedBEns-talpini2025fedbensoneshotfederatedlearning}.
By fully exploiting the rich multimodal interactions inherent in VLMs, our method obviates the need for model training on both the server and client sides.

\section{Proofs and Derivations in Section~\ref{method}}

In this section, we present the \textbf{the proof for Lemma 1} in subsection B.1, and \textbf{the proof for Theorem 1} in subsection B.2, along with \textbf{privacy analysis} in subsection B.3.

\subsection{Background and Proof for Lemma~\ref{lemma-1} with Further Derivations}
\textbf{Wishart distribution.} The Wishart distribution $\mathcal{W}(\mathbf{S},\nu)$ is the probability distribution of the sample covariance matrix of a multivariate normal distribution’s covariance matrix, used to describe the statistical properties of positive definite matrices. 
Suppose the random variable $\Lambda \sim \mathcal{W}(\mathbf{S},\nu) \in \mathcal{R}^{d\times d}$, the probability density distribution  is defined as follows:
\begin{equation*}
\begin{split}
\mathcal{W}(\boldsymbol{\Lambda} \mid \mathbf{S}, \nu)=\frac{1}{Z_{W}}|\boldsymbol{\Lambda}|^{(\nu-D-1) / 2} \exp \left(-\frac{1}{2} \operatorname{tr}\left(\boldsymbol{\Lambda} \mathbf{S}^{-1}\right)\right)
\end{split}
\end{equation*}
where $Z_{\mathcal{W}}$ is the probabilistic regularization constant.

\noindent \textbf{Inverse Wishart distribution.} The inverse Wishart distribution $\mathcal{IW}(\mathbf{S},\nu)$, which is defined as the inverse of Wishart distribution ($\mathbf{\Lambda} \sim \mathcal{W}(\mathbf{S},\nu)$, then $\mathbf{\Lambda}^{-1} \sim \mathcal{IW}(\mathbf{S},\nu)$), is another probability distribution defined on real-valued positive-definite matrices. 
Suppose the random variable $\Sigma \sim \mathcal{IW}(\mathbf{S},\nu) \in \mathcal{R}^{d\times d}$, the probability density distribution  is defined as follows:
\begin{equation*}
\begin{split}
\mathcal{IW}(\boldsymbol{\Sigma} \mid \mathbf{S}, \nu) =\frac{1}{Z_{IW}}|\boldsymbol{\Sigma}|^{-(\nu+D+1) / 2} \exp \left(-\frac{1}{2} \operatorname{tr}\left(\mathbf{S} \boldsymbol{\Sigma}^{-1}\right)\right) \\
\end{split}
\end{equation*}
where $Z_{\mathcal{IW}}$ is the probabilistic regularization constant. The Wishart distribution and inverse Wishart distribution have been defined in various forms in previous literature, and here we adopt the definition from~\cite{prml-bishop2006pattern} for consistency.

\noindent \textbf{Proof for Lemma~\ref{lemma-1}.} We first prove the inference results of the hierarchical Bayesian model as mentioned in section~\ref{method}.
To guarantee both completeness and coherence, we revisit the problem setting here.

Given a Gaussian mixture model over visual representation $\mathbf{z}$, whose density $\mathbb{P}(\mathbf{z})$ is given by $\mathbb{P}(\mathbf{\mathbf{z}})=\sum_{i=1}^C p_i \mathcal{N}\left(\mathbf{\mathbf{z}} ; \mathbf{w}_i, \Sigma\right)$, where the density $\mathcal{N}\left(\mathbf{z} ; \mathbf{w}_c, \Sigma\right)$ is 
\begin{equation*} \label{gaussian-appendix}
\begin{split}
\frac{1}{\sqrt{(2 \pi)^d|\Sigma|}} \exp \left\{-\frac{1}{2}\left(\mathbf{z}-\mathbf{w}_c\right)^{\top} \Sigma^{-1}\left(\mathbf{z}-\mathbf{w}_c\right)\right\}.
\end{split}
\end{equation*}
The probability $p_i = p(y=i):= 1/C, i = 1, 2, ..., C$ is the prior probability of the corresponding class, which is set to be uniform here.

Therefore, the parameters that remain to be inferred are the means $(\mathbf{w}_{1},\dots,\mathbf{w}_{C})$ and the covariance matrix $\Sigma$ in the Gaussian mixture model.

We introduce Bayesian statistics to characterize the variation of parameters as the data increase.
First, it is necessary to introduce a prior distribution on unknown parameters.
It is reasonable to assume that given covariance $\Sigma$, the means of each class are independent of each other, namely $p(\mathbf{w}_{1},\dots,\mathbf{w}_{C},\Sigma)=p(\Sigma)\Pi_{c=1}^C p(\mathbf{w}_c| \Sigma)$.
For prompt distributions over global visual representation, the Bayesian inference is constructed as follows
\begin{equation*}
\begin{split}
    &\pi(\mathbf{w}_{1},\dots,\mathbf{w}_{C},\Sigma \mid D)\\
    & \quad \propto L(D \mid \mathbf{w}_{1},\dots,\mathbf{w}_{C},\Sigma) \pi(\mathbf{w}_{1},\dots,\mathbf{w}_{C},\Sigma),\\
\end{split}
\end{equation*}
where $\pi$ is the prior distribution over $(\mathbf{w}_{1},\dots,\mathbf{w}_{C},\Sigma)$ and the likelihood function $L(D\mid \mathbf{w}_{1},\dots,\mathbf{w}_{C},\Sigma)$ is formulated as
\begin{flalign*}\small
\begin{split}
    &L(D \mid \mathbf{w}_{1}, \dots,\mathbf{w}_{C},\Sigma) := \prod_{(\mathbf{x},y) \in D}\prod_{c \in [C]}e^{\mathbb{I}(y=c) \log \mathcal{N}(\mathbf{z};\mathbf{w}_c, \Sigma)}\\
    &=\prod_{c \in [C]}(2 \pi)^{-\frac{N_c d}{2}}|\Sigma|^{-\frac{N_c}{2}} \times \\
    & \qquad \exp \left(-\frac{1}{2} \sum_{i=1}^{N_c}\left(\mathbf{z}_{c,i}-\mathbf{w}_c\right)^\top \boldsymbol{\Sigma}^{-1}\left(\mathbf{z}_{c,i}-\mathbf{w}_c\right)\right)\\
    &=(2 \pi)^{-\frac{N d}{2}}|\Sigma|^{-\frac{N}{2}} \times \\ 
    & \qquad \exp \left(-\frac{1}{2}\sum_{c=1}^C \sum_{i=1}^{N_c}\left(\mathbf{z}_{c,i}-\mathbf{w}_c\right)^\top \boldsymbol{\Sigma}^{-1}\left(\mathbf{z}_{c,i}-\mathbf{w}_c\right)\right)\\
    &=(2 \pi)^{-\frac{N d}{2}}|\Sigma|^{-\frac{N}{2}} \times \\ 
    & \qquad \exp \left(-\frac{1}{2}\operatorname{Tr}\left[\boldsymbol{\Sigma}^{-1}\sum_{c=1}^C \sum_{i=1}^{N_c}\left(\mathbf{z}_{c,i}-\mathbf{w}_c\right) \left(\mathbf{z}_{c,i}-\mathbf{w}_c\right)^\top\right]\right)\\
\end{split}
\end{flalign*}
where $\mathbf{z}_{c,i}:=f(\mathbf{x_{c,i}})$ and $\mathbf{x_{c,i}}$ denote the $i$-th data for the $c$-th class in the global view.

In order to develop a training-free adaptation method, we introduce a conjugate prior distribution for this Bayesian inference, ensuring that the posterior distribution has a closed-form expression that is computationally feasible.

Given the prior distribution $\pi$ over the global parameter $(\mathbf{w}_{1:C},\Sigma)$, assume that the family of distributions satisfies
\begin{equation*}
\begin{split}
\Sigma \sim \mathcal{I W}(\Sigma ; \mathbf{S}_0, \nu_0), \quad \mathbf{w}_c| \Sigma \sim \mathcal{N}(\mathbf{x} ; \mathbf{m}_{0,c}, \frac{1}{\kappa_{0,c}} \Sigma), \ c\in [C],
\end{split}
\end{equation*}
where $\mathbf{S}_0 \in \mathbb{R}^{d\times d}$ is positive definite, $\nu_0 \in \mathbb{R}$, and $\mathbf{m}_{0,c} \in \mathbb{R}^{d}, \kappa_{0,c} \in \mathbb{R}_+,c\in [C]$ are the parameters of the prior distribution. 
The prior can be expressed as
\begin{flalign*}\small
\begin{split}
&\pi(\mathbf{w}_{1},\dots,\mathbf{w}_{C},\Sigma)=p(\Sigma)\Pi_{c=1}^C p(\mathbf{w}_c| \Sigma)\\
&=\prod_{c=1}^C (2 \pi)^{-d / 2}|\kappa_{0,c}\boldsymbol{\Sigma}|^{-\frac{1}{2}} \exp \left(-\frac{\kappa_{0,c}}{2} \left(\mathbf{w}_c-\mathbf{m}_{0,c}\right)^\top \boldsymbol{\Sigma}^{-1}\left(\mathbf{w}_c-\mathbf{m}_{0,c}\right)\right)\\
& \qquad \times \frac{1}{Z_{I W}}|\boldsymbol{\Sigma}|^{-(\nu_0+d+1) / 2} \exp \left(-\frac{1}{2} \operatorname{Tr}\left(\mathbf{S_0} \boldsymbol{\Sigma}^{-1}\right)\right)\\
&= \frac{1}{Z_{I W}}(2 \pi)^{-C d / 2} |\Sigma|^{-\frac{(\nu_0+d+C+1)}{2}} \prod_{c=1}^C|\kappa_{0,c}|^{-\frac{1}{2}}\\
& \quad \times\exp \left(-\frac{1}{2} \operatorname{Tr} \left[ \Sigma^{-1} \sum_{c=1}^C\kappa_{0,c}\left(\mathbf{w}_c-\mathbf{m}_{0,c}\right)\left(\mathbf{w}_c-\mathbf{m}_{0,c}\right)^{\top} +\mathbf{S_0} \boldsymbol{\Sigma}^{-1} \right] \right).
\end{split}
\end{flalign*}
Hence, we have 
\begin{flalign*}
\small
\begin{split}
&\pi(\mathbf{w}_{1},\dots,\mathbf{w}_{C},\Sigma)\\
&\propto L(D\mid \mathbf{w}_{1},\dots,\mathbf{w}_{C},\Sigma) \pi(\mathbf{w}_{1},\dots,\mathbf{w}_{C},\Sigma\mid D)\\
& \propto |\Sigma|^{-\frac{(\nu_0+d+N+C+1)}{2}}\\
& \ \ \times\exp \Bigg(-\frac{1}{2} \operatorname{Tr} \bigg[ \boldsymbol{\Sigma}^{-1} \sum_{c=1}^C \kappa_{0,c} \left(\mathbf{w}_c-\mathbf{m}_{0,c}\right)\left(\mathbf{w}_c-\mathbf{m}_{0,c}\right)^{\top} +\mathbf{S_0} \boldsymbol{\Sigma}^{-1} \bigg] \Bigg)\\
&\ \ \times \exp \left(-\frac{1}{2}\operatorname{Tr}\left[\boldsymbol{\Sigma}^{-1}\sum_{c=1}^C \sum_{i=1}^{N_c}\left(\mathbf{z}_{c,i}-\mathbf{w}_c\right) \left(\mathbf{z}_{c,i}-\mathbf{w}_c\right)^\top\right]\right)\\
& \propto |\Sigma|^{-\frac{(\nu_0+d+N+C+1)}{2}}\\
& \ \ \times\exp \Bigg(\\
& \ \quad -\frac{1}{2} \operatorname{Tr} \bigg[ \boldsymbol{\Sigma}^{-1} \sum_{c=1}^C \kappa_{0,c} \big(\mathbf{w}_c\mathbf{w}_c^\top-2 \mathbf{w}_c \mathbf{m}_{0,c}^\top + \mathbf{m}_{0,c} \mathbf{m}_{0,c}^\top\big)+\boldsymbol{\Sigma}^{-1}\mathbf{S_0} \bigg] \Bigg)\\
&\ \ \times \exp \left(-\frac{1}{2}\operatorname{Tr}\left[\boldsymbol{\Sigma}^{-1}\sum_{c=1}^C N_c \mathbf{w}_c \mathbf{w}_c^\top 
 -2\mathbf{w}_c N_c \bar{\mathbf{z}}_c + \sum_{i=1}^{N_c} \mathbf{z}_{c,i} \mathbf{z}_{c,i}^\top \right]\right)\\
& \propto |\Sigma|^{-\frac{(\nu_0+d+N+C+1)}{2}}\\
& \ \ \times\exp \Bigg(-\frac{1}{2} \operatorname{Tr} \Bigg[ \boldsymbol{\Sigma}^{-1} \sum_{c=1}^C \Bigg((N_C + \kappa_{0,c})\bigg(\mathbf{w}_c - \frac{\kappa_{0,c} \mathbf{m}_{0,c}+N_c \bar{\mathbf{z}}_c}{N_c+\kappa_{0,c}}\bigg)\\
& \ \ \quad \times\bigg(\mathbf{w}_c - \frac{\kappa_{0,c} \mathbf{m}_{0,c}+N_c \bar{\mathbf{z}}_c}{N_c+\kappa_{0,c}}\bigg)^\top + \frac{N_c\kappa_{0,c}}{N_c+\kappa_{0,c}} \big(\bar{\mathbf{z}}_c-\mathbf{m}_{0,c}\big)\big(\bar{\mathbf{z}}_c-\mathbf{m}_{0,c}\big)^\top\\
&\ \ \quad +\sum_{i=1}^{N_c} \big(\mathbf{z}_{c,i}-\bar{\mathbf{z}}_c\big) \big(\mathbf{z}_{c,i}-\bar{\mathbf{z}}_c\big)^\top\Bigg) +\boldsymbol{\Sigma}^{-1}\mathbf{S_0}\Bigg] \Bigg).\\
\end{split}
\end{flalign*}
Clearly, the posterior distribution shares the same functional form as the prior, indicating that the above prior is conjugate.
The posterior can be shown to be the prior $\pi(\mathbf{w}_{1},\dots,\mathbf{w}_{C},\Sigma)$ with updated parameters
\begin{flalign*}\small
\begin{split}
&\pi(\mathbf{w}_{1},\dots,\mathbf{w}_{C},\Sigma \mid D) \propto |\Sigma|^{-\frac{(\nu_q+d+C+1)}{2}}\\
& \quad \times\exp \Bigg(-\frac{1}{2} \operatorname{Tr} \Bigg[ \boldsymbol{\Sigma}^{-1} \sum_{c=1}^C \kappa_{q,c}\big(\mathbf{w}_c - \mathbf{m}_{q,c}\big)\big(\mathbf{w}_c - \mathbf{m}_{q,c}\big)^\top +\Sigma^{-1}\mathbf{S}_q\Bigg] \Bigg).\\
\end{split}
\end{flalign*}
\begin{flalign}\small\label{posterior-formula}
\begin{split}
&\nu_q = \nu_0 + N\\
&\kappa_{q,c} = \kappa_{0,c}+ N_c\\
&\mathbf{m}_{q,c} = \frac{\kappa_{0,c} \mathbf{m}_{0,c}+N_c \bar{\mathbf{z}}_c}{N_c+\kappa_{0,c}}\\
&\mathbf{S}_q \ \ \ =\mathbf{S}_0+\sum_{c \in [C]}\mathbf{S}_c (\bar{\mathbf{z}}_c) + \sum_{c \in [C]}\frac{N_c \kappa_0}{N_c+\kappa_0} (\bar{\mathbf{z}}_c-\mathbf{m}_{0,c})(\bar{\mathbf{z}}_c-\mathbf{m}_{0,c})^\top \\
&\quad \ \ \ \ = \mathbf{S}_0+\sum_{c \in [C]}\mathbf{S}_c  + \sum_{c \in [C]}\big(\kappa_{0.c} \mathbf{m}_{0,c}\mathbf{m}_{0,c}^\top - \kappa_{g,c} \mathbf{m}_{q,c}\mathbf{m}_{q,c}\big), \\
\end{split}
\end{flalign}
where $\mathbf{S}_c (\bar{\mathbf{z}}_c):=\sum_{i=1}^{N_c} \big(\mathbf{z}_{c,i}-\bar{\mathbf{z}}_c\big) \big(\mathbf{z}_{c,i}-\bar{\mathbf{z}}_c\big)^\top, \mathbf{S}_c:=\sum_{i=1}^{N_c} \mathbf{z}_{c,i}\mathbf{z}_{c,i}^\top, $

\hfill \ensuremath{\square}

\noindent \textbf{Derivations for global and personalized posteriors.}
In performing posterior inference on the global prompt distribution (\textbf{Step 1}), we use an uninformative prior: $\mathbf{S}_0=s_0 \mathbf{I}, \mathbf{m}_{c,0}=\mathbf{0}, \nu_0=0$ and $s_0 \approx \kappa_{c,0} \approx0$ as some small positive number.
By plugging the uninformative prior into~\eqref{posterior-formula} , we arrive at the global posterior for prompt prototype distributions,
\begin{equation} \label{global-posterior}
\begin{split}
    \kappa_{g,c} & = N_c, \qquad \quad \mathbf{m}_{g,c} = \overline{\mathbf{z}}_c:= \frac{1}{N_c}\sum_{i=1}^{N_c}\mathbf{z}_{c,i}, \\
    \nu_g &=\sum_{c \in [C]}N_c, \quad
    \mathbf{S}_g = \sum_{c \in [C]}\sum_{i=1}^{N_c}(\mathbf{z}_{c,i} - \overline{\mathbf{z}}_c)(\mathbf{z}_{c,i} - \overline{\mathbf{z}}_c)^\top.\\
\end{split}
\end{equation}
For the inference of personalized prompt prototype distributions (\textbf{Step 3}), we take the global posterior $(\mathbf{S}_g,\nu_g ,\mathbf{m}_{g,c},\kappa_{g,c})$ as the prior and integrate local evidence to derive the personalized prompt distribution over visual representations.
For $k$-th local client, with the local power prior method, the global likelihood turns to 
\begin{flalign*}\small
\begin{split}
    &L(D \mid \mathbf{w}_{1}, \dots,\mathbf{w}_{C},\Sigma)^{a_k} = \prod_{(\mathbf{x},y) \in D}\prod_{c \in [C]}e^{a_k \mathbb{I}(y=c) \log \mathcal{N}(\mathbf{z};\mathbf{w}_c, \Sigma)}\\
    &=(2 \pi)^{-\frac{N d}{2}}|\Sigma|^{-\frac{N}{2}} \times\\
    &\quad \exp \left(-\frac{1}{2}\operatorname{Tr}\left[\boldsymbol{\Sigma}^{-1}\sum_{c=1}^C a_k\sum_{i=1}^{N_c}\left(\mathbf{z}_{c,i}-\mathbf{w}_c\right) \left(\mathbf{z}_{c,i}-\mathbf{w}_c\right)^\top\right]\right).\\
\end{split}
\end{flalign*}
Substituting it in the conjugate posterior~\eqref{posterior-formula}, we get
\begin{flalign} \small \label{local-posterior-lemma}
\begin{split}
&\nu'_q = \nu_0 + \alpha N\\
&\kappa'_{q,c} = \kappa_{0,c} + \alpha N_c\\
&\mathbf{m}'_{q,c} = \frac{\kappa_{0,c} \mathbf{m}_{0,c}+\alpha N_c \bar{\mathbf{z}}_c}{\alpha N_c+\kappa_{0,c}}\\
&\mathbf{S}'_q =\mathbf{S}_0+\alpha\sum_{c \in [C]}\mathbf{S}_c (\bar{\mathbf{z}}_c) + \sum_{c \in [C]}\frac{\alpha N_c \kappa_0}{N_c+\kappa_0} (\bar{\mathbf{z}}_c-\mathbf{m}_{0,c})(\bar{\mathbf{z}}_c-\mathbf{m}_{0,c})^\top \\
&\quad= \mathbf{S}_0+\alpha\sum_{c \in [C]}\mathbf{S}_c  + \sum_{c \in [C]}\big(\kappa_{0.c} \mathbf{m}_{0,c}\mathbf{m}_{0,c}^\top - \kappa_{g,c} \mathbf{m}_{q,c}\mathbf{m}_{q,c}\big). \\
\end{split}
\end{flalign}
Utilize the global posterior $(\mathbf{S}'_g,\nu'_g ,\mathbf{m}'_{g,c},\kappa'_{g,c})$ with an uninformative prior. according to~\eqref{local-bayes} and~\eqref{local-posterior-lemma}, the local Bayesian posterior can be derived as 
\begin{flalign}
\small \label{posterior-formula-alpha-appendix}
\begin{split}
&\nu^k_l =  \alpha N + N^k\\
&\kappa^k_{l,c} = \alpha N_c + N^k_c\\
&\mathbf{m}^k_{l,c} = \frac{\alpha N_c \bar{\mathbf{z}}_c + N^k_c \bar{\mathbf{z}}^k_c}{\alpha N_c+N^k_c}\\
&\mathbf{S}^k_l \ \ \ =\alpha\sum_{c \in [C]}\mathbf{S}_c + \sum_{c \in [C]}\mathbf{S}_c^k + \sum_{c \in [C]}\big(\alpha N_c \bar{\mathbf{z}}_c \bar{\mathbf{z}}_c^\top - \kappa^k_{l,c} \mathbf{m}^k_{l,c}\mathbf{m}^k_{l,c}\big)\\
\end{split}
\end{flalign}
Thus, we obtain the personalized posterior $(\mathbf{S}^k_l,\nu^k_l ,\mathbf{m}^k_{l,c},\kappa^k_{l,c})$ inference in~\eqref{posterior-formula-alpha-appendix}, resulting in an estimated personalized prompt distribution for visual representations.

\subsection{Proof for Theorem 1}


The proof of Theorem 1 needs the following lemma.

\begin{lemma} \cite{ml-textbook-understand}
    Let $\mathcal{H}$ be a class of binary functions from $\mathcal{X}$ to $\{-1,1\}$ and i.i.d. samples $x_1, \ldots, x_N \sim \mathbb{P}_X$. Then with probability at least $1-\delta$ over the samples, $\forall f \in \mathcal{H}$,
$$
\left|\mathcal{R}(f)-\widehat{\mathcal{R}}(f)\right| \leq B' \sqrt{\frac{d_{\mathrm{VC}}(\mathcal{H})+\ln 1 / \delta}{N}}
$$
where the generalization error $\mathcal{R}(f):= \mathbb{E}_{(x,y)\sim \mathcal{D}}[\ell(f(x),y)]$ and the empirical error $\widehat{\mathcal{R}}(f):= \frac{1}{N}\sum_{i=1}^N\ell(f(x_i), y_i)$, for some universal constant $B' >0$.
\end{lemma}

\noindent \textbf{Proof of Theorem 1.}
Considering $\ell$ to be the convex logistic loss function (e.g., cross-entropy loss) applied to binary classification tasks, we have

\begin{align} \label{proof-eq-1}
    \ell(f(\mathbf{z}), y)&=\ell\left(\eta(\mathbf{z})f_1(\mathbf{z}) + (1-\eta(\mathbf{z}))f_2(\mathbf{z}), y \right) \notag\\
    &\leq  \eta(\mathbf{z})\ell(f_1(\mathbf{z}),y) + (1-\eta(\mathbf{z})) \ell\left(f_2(\mathbf{z}),y)\right)
\end{align}

When computing the expectation of Equation~\eqref{proof-eq-1} and leveraging the properties of expectation, the above equation is also satisfied. 
The definition of covariance tells us that
\begin{align} \label{proof-eq-2}
\operatorname{Cov}(X,Y) &:= \mathbb{E}\left[(X-\mathbb{E}[X])(Y-\mathbb{E}[Y])\right] \notag\\
& = \mathbb{E}[XY]-\mathbb{E}[X]\mathbb{E}[Y].
\end{align}
To simplify notation, we denote $\mathbb{E}_{(\mathbf{z},y)\sim \mathcal{D}}$ as $\mathbb{E}$ in subsequent derivations.

\begin{align*} \label{proof-eq-3}
    \mathcal{R}(f) &= \mathbb{E}[\ell(f(x),y)]\\
    & \leq \mathbb{E}_{(\mathbf{z},y)\sim \mathcal{D}} \left[ \eta(\mathbf{z})\ell(f_1(\mathbf{z}),y) + (1-\eta(\mathbf{z})) \ell\left(f_2(\mathbf{z}),y)\right)\right]\\
    & = \mathbb{E}[\eta(\mathbf{z})]\mathbb{E}[\ell(f_1(\mathbf{z}),y)] + \operatorname{Cov}\left( \eta(\mathbf{z}),\ell(f_1(\mathbf{z}),y) \right)\\
    & \quad + \mathbb{E}[1 - \eta(\mathbf{z})]\mathbb{E}[\ell(f_2(\mathbf{z}),y)] - \operatorname{Cov}\left( \eta(\mathbf{z}),\ell(f_2(\mathbf{z}),y) \right)\\
    & \leq B \sqrt{\frac{d_{\mathrm{VC}}(\mathcal{H})+\ln 1 / \delta}{N}} + \widehat{\mathcal{R}}(f_1) + \widehat{\mathcal{R}}(f_2)\\
    & \quad + \operatorname{Cov}\left( \eta(\mathbf{z}),\ell(f_1(\mathbf{z}),y) - \ell(f_2(\mathbf{z}),y) \right)
\end{align*}
The second inequality is due to the convexity of loss function as depicted in~\eqref{proof-eq-1}. The third equality is derived from the definition of covariance in~\eqref{proof-eq-2}. the The last inequality is by taking a union bound in Lemma 2, and setting $B=4B'$. Thus, the proof is complete. \hfill \ensuremath{\square}

\subsection{Privacy Analysis}\label{privacy}

This subsection focuses on the privacy aspects of the \M~framework.
Client–server interactions in our approach are conducted through visual and language pipelines, as depicted in Figure~\ref{framework-fig-1}.

\noindent \textbf{Visual pipeline.} In the visual pipeline, the primary client–server interaction aims to estimate the global prompt distribution, especially $\mathbf{m}_{g,c}$ and $\mathbf{S}_g$ in~\eqref{global-bayes}.
These parameters can be directly calculated with each client's local data statistics:
\begin{equation*} \label{federated-setting-estimation}
\small
\begin{split}
&\mathbf{m}_{g,c}=\overline{\mathbf{z}}_c=\frac{1}{\sum_{k \in [K]} N_{c}^k} \sum_{k \in [K]} \sum_{i=1}^{N^k_c}\mathbf{z}_{c,i}^k,\\
&\mathbf{S}_g=\sum_{k\in [K]} \sum_{c \in [C]}A_{k,c}, \\
& A_{k,c}:=\Bigg[\sum_{i=1}^{N_c^k}(\mathbf{z}_{c,i}^k-\overline{\mathbf{z}}_c^k)(\mathbf{z}_{c,i}^k-\overline{\mathbf{z}}_c^k)^\top - N_c^k(\overline{\mathbf{z}}_c^k - \overline{\mathbf{z}}_c)(\overline{\mathbf{z}}_c^k - \overline{\mathbf{z}}_c)^\top\Bigg].\\ 
\end{split}
\end{equation*}
Note that only local mean and variance statistics are exchanged, from which an adversary cannot infer individual user data.
Moreover, if clients prefer not to reveal these statistics, they can establish encrypted federated computation methods, such as SecAgg~\cite{SecAgg-10.1145/3133956.3133982}.

%
\noindent \textbf{Textual pipeline.} In the textual pipeline, to ensure security, we encode text prompts into embeddings using the CLIP text encoder and transmit them to the server for subsequent processing. Such embeddings have been widely regarded as privacy-safe for exchange in prior prompt learning work~\cite{promptfl-guo2023promptfl,fedprompt-iccsp-10095356,pfedprompt-guo2023pfedprompt,fedopt-prompt-li2024global,portfolio-fed-theory-pan2024federated}. Moreover, there is no requirement that local LLM architectures or versions be identical.
In terms of communication, analogous to the visual modality, text importance scores also will not expose individual user data on one-shot condition, and privacy can be further enhanced by employing encrypted aggregation protocols~\cite{SecAgg-10.1145/3133956.3133982,fed-privacy-server-bonawitz2019towards,Privacy-preserving-aggregation-servey-liu2022privacy}.

\section{Additional Experimental Setting}\label{datasets-and-experimental-setting-appendix}
\subsection{Datasets and Data Heterogeneity}
We assess the effectiveness of the proposed \M~across nine publicly available benchmark datasets under various federated configurations to simulate different types of data heterogeneity.
Following prior works on adapting VLMs in the context of FL~\cite{portfolio-fed-theory-pan2024federated,DP-FPL-tran2025privacypreserving,pfedprompt-guo2023pfedprompt,promptfl-guo2023promptfl,fedopt-prompt-li2024global} , we evaluate \M’s few-shot performance using five representative visual classification datasets commonly employed to benchmark CLIP~\cite{CLIP}: \textbf{OxfordPets}~\cite{OxfordPets-parkhi2012cats}, \textbf{Flowers102}~\cite{Flowers102-nilsback2008automated}, \textbf{DTD}~\cite{DTD-cimpoi2014describing}, \textbf{Caltech101}~\cite{Caltech101-fei2004learning}, and \textbf{Food101}~\cite{Food101-bossard2014food}, hereafter referred to collectively as the CLIP datasets.
In addition, we also employ two standard image benchmarks, \textbf{CIFAR‑10} and \textbf{CIFAR‑100}~\cite{cifar-gong2012geodesic}.
We simulate label shift by partitioning datasets among clients by sampling from a symmetric Dirichlet distribution with $\beta=0.3$, following the approach of~\cite{Dirichlet-partition-1-cao2023knowledge,Dirichlet-partition-2-hsu2019measuring}
Consider another critical type of data heterogeneity in FL, feature shift~\cite{fedBN-li2021fedbn}, we utilize two multi-domain datasets: \textbf{DomainNet}~\cite{domainnet-peng2019moment} (six domains), and \textbf{Office‑Caltech10}~\cite{cifar-gong2012geodesic} (four domains).
Consistent with prior work~\cite{fedBN-li2021fedbn,fedpgp-harmon-pmlr-v235-cui24c}, each client in the federated system is assigned data from a single, distinct domain.

\subsection{Baselines}
Since \textit{no previous work was found that investigates training-free and one-shot distributed adaptation method for VLMs} within a FL framework, we compare the performance of \M~with four categories of baselines: 
(1) Four existing prompt learning federated learning methods:  PromptFolio~\cite{portfolio-fed-theory-pan2024federated} introduces a global–local prompt portfolio that fuses task‑relevant and personalized features; DP-PFL~\cite{DP-FPL-tran2025privacypreserving} presents a privacy‑preserving federated prompt learning model that factorizes local prompts with a residual term to enable personalized adaptation; PromptFL~\cite{promptfl-guo2023promptfl} learns a unified prompt across clients; pFedPrompt~\cite{pfedprompt-guo2023pfedprompt} learns a shared prompt with personalized visual attention modules; 
(2) Three local adapting methods: Zero-shot CLIP~\cite{CLIP} with hand-crafted text prompt templates; Wang et al.~\cite{GDA-CLIP-wang2024hard} apply Gaussian Discriminant Analysis (GDA) to enhance the downstream training-free classification of CLIP, denoted as CLIP-GDA in our experiments; CoOp~\cite{COOP-zhou2022learning} uses learnable prompt vectors to extract effective text modal information, trained locally on each client.
(3) Three adapted methods derived from advanced one-shot techniques that operate solely with client-side training resources, combined with the backbone prompt learning method~\cite{promptfl-guo2023promptfl} in FL: FedLPA~\cite{oneshot-fedlpa-liu2024fedlpa} + PromptFL, FENS~\cite{FENS-allouah2024revisiting} + PromptFL, FedFisher~\cite{fedfisher-oneshot-pmlr-v238-jhunjhunwala24a} + PromptFL.
(4) FedAvg~\cite{fedavg}. Since existing traditional one-shot FL methods are still focused on matching the accuracy of multi-round iterative training methods like FedAvg, we use FedAvg as an alternative to traditional one-shot FL baselines.

\subsection{Implementation Details}
\noindent \textbf{Data partition.}

\noindent(1) \textbf{CLIP datasets.} Each dataset in CLIP datasets is partitioned into $N = 10$ clients, defaulting if not explicitly specified, each with a disjoint set of classes evenly and randomly assigned to the clients.

\noindent(2) \textbf{CIFAR10 and CIFAR100.} We split $N = 100$ clients resulting from $Dir(\beta = 0.3)$ partition.

\noindent(3) \textbf{DomainNet and Office-Caltech10.} Each client in the federated system is assigned data from a single unique domain. 

\noindent \textbf{Performance evalution.}
We present the results using two representative backbones, ResNet50~\cite{resnet-he2016deep} and ViT-B16~\cite{vit-b16-dosovitskiy2020image}, defaulting to ViT-B16 if not explicitly specified.
For other hyperparameters, such as learning rate and local epochs in the aforementioned baselines, we adhere to the original configurations from these studies.
Given that the results of our model are deterministic due to its training-free nature, we do not present results with statistical variations, which is typical in zero-shot or training-free studies~\cite{Frolic-zhu2024enhancing-ProDA,awt-zhu2024awt-nips2024,GDA-CLIP-wang2024hard}.
We measure performance by computing the global accuracy across each client’s private test set drawn from the same distribution as its training data.

\noindent \textbf{Textual augmentation.} In the textual pipeline, we employ LLMs to generate prompts that enhance CLIP’s classification performance. Specifically, for the CLIP datasets and Office–Caltech10, we refer to Table~13 in~\cite{awt-zhu2024awt-nips2024}. For DomainNet, we use "an image dataset that comprises common object images across six distinct domains: real photos, paintings, clip art, quick sketches, infographics, and sketches. For a total of approximately 600,000 images spanning 345 categories". For the CIFAR dataset, we do not employ any textual augmentation. For each dataset, the prompts below are used to produce broad feature descriptions of every category:

\noindent \textit{Please write 3–4 sentences (each less than 15 words) describing a typical visual characterization of \{category\}.  
Each sentence must start with “The \{category\} ” followed exactly one visual attribute (e.g., shape, color, texture, pattern, size).}

Subsequently, the feature descriptions are converted into embeddings by the CLIP text encoder and transmitted to the server. Given the potential heterogeneity in the LLMs (architecture or version) deployed on each client, we apply the following filtering and aggregation methods to perform an initial screening and processing of these text embeddings, after which further alignment is carried out by the techniques within our textual pipeline. To find text prompts that remain robust for every client, we begin by screening text embeddings $e_{c, a}^k$ that meet the following conditions:

$$
\frac{\left\langle e_{c, a}^k, e_{c, b}^{k'}\right\rangle}{\left\|e_{c, a}^k\right\|\left\|e_{c, b}^{k'}\right\|}<\kappa,
$$
for ${k'} \neq k, {k'} \in [K]$. We then average the selected text-embedding clusters from each client and feed the result into the next component of the textual pipeline.

\end{document}